\documentclass[fleqn,11pt]{wlscirep}
\usepackage[utf8]{inputenc}
\usepackage[T1]{fontenc}
\usepackage{bm}
\usepackage{float}
\usepackage{soul}
\graphicspath{{./pix/}}
\usepackage{subfig}
\usepackage{algorithm}
\usepackage{algorithmic}

\DeclareMathAlphabet{\mathcal}{OMS}{cmsy}{m}{n}

\title{TLab: Traffic Map Movie Forecasting Based on HR-NET} 
\author[1,a,*]{Fanyou Wu}
\author[2,b,*]{Yang Liu}
\author[2]{Zhiyuan Liu}
\author[3]{Xiaobo Qu}
\author[1]{Rado Gazo}
\author[1]{Eva Haviarova}
\affil[1]{Purdue University, Department of Forestry and Natural Resource, West Lafayette, USA }
\affil[2]{Southeast Univeristy, School of Transportation, Nanjing, China}
\affil[3]{Chalmers University of Technology, Department of Architecture and Civil Engineering, Gothenburg, Sweden}
\affil[a]{Email: wu1297@purdue.edu}
\affil[b]{Email: 230179629@seu.edu.cn}
\affil[*]{Equal Contribution and Communication Authors}

\begin{abstract}
The problem of the effective prediction for large-scale spatio-temporal traffic data has long haunted researchers in the field of intelligent transportation. Limited by the quantity of data, citywide traffic state prediction was seldom achieved. Hence the complex urban transportation system of an entire city cannot be truly understood. Thanks to the efforts of organizations like IARAI, the massive open data provided by them has made the research possible. In our 2020 Competition solution, we further design multiple variants based on HR-NET and UNet. Through feature engineering, the hand-crafted features are input into the model in a form of channels. It is worth noting that, to learn the inherent attributes of geographical locations, we proposed a novel method called geo-embedding, which contributes to significant improvement in the accuracy of the model. In addition, we explored the influence of the selection of activation functions and optimizers, as well as tricks during model training on the model performance. In terms of prediction accuracy, our solution has won 2nd place in NeurIPS 2020, Traffic4cast Challenge.
The source codes are available in \url{https://github.com/wufanyou/Traffic4Cast-2020-TLab}.

\end{abstract}
\begin{document}

\flushbottom
\maketitle
\thispagestyle{empty}
\section{Introduction}

Real-time traffic state prediction is an essential component for traffic control and management in an urban road network. The ability to predict future traffic state (e.g., flow, speed) can help improve traffic conditions, fleet organization, utilization rate, and social welfare~\cite{ke2017short,yao2018deep}.
Essentially, the traffic prediction is a time series problem, which is performed based on the changes in historical demand. A representative time-series prediction tool is the recurrent neural network (RNN), along with its diverse variants~\cite{liu2019deeppf,fu2016using}. Apart from the temporal dimension, the correlation in the spatial dimension is also extensively incorporated by many works. Regions that are close to each other or share similar land-use structures may exhibit a homogeneous demand pattern. Techniques widely applied in computer vision like convolutional neural network (CNN)~\cite{yao2018deep,zhang2017deep} and the emerging graph-based networks~\cite{geng2019spatiotemporal,li2017diffusion,Pan:2019:UTP:3292500.3330884,yu2018spatio} are often adopted. Furthermore, multi-source data are also introduced in some literature to allow for the external influencing factors, such as weather conditions and neighboring points-of-interest~\cite{koesdwiady2016improving,liao2018deep}.

\section{Data Description and Problem Definition}

The organizer provides industrial-scale, real-world data for 3 entire cities Berlin, Istanbul and Moscow over a year period~\cite{pmlr-v123-kreil20a}. The organizer divides each city into a $436\times495$ grid; each pixel of this grid represents a region of $100\,m \times 100\,m$. In this competition, the dataset is comprised of dynamic data (\textit{e.g.}, traffic speed), and static data (\textit{e.g.}, the number of entertainment amenities). In the dynamic dataset, there are a total of 288 frames each day, each frame representing the aggregated information with nine channels over five minutes, including the traffic volume, speed, and an aggregated incident level channel (a higher value indicates a more severe incident) in each ordinal direction (\textit{e.g.}, NW, NE, SW, SE). The static dataset describes the locations of road junctions and points of interest, such as food and drink, shopping, parking, transit, etc.  %The sample images are shown in \hyperref[traffic4cast:samples]{Figure}~\ref{traffic4cast:samples}. 
The data of volume,speed, and incident level are scaled to $[0,255]$ through a min-max scaler. Missing values are represented by 0.
\vspace{1ex}

\textbf{Problem:} This challenge is a multi-task learning problem, \textit{i.e.}, use the given historical data to predict trafﬁc volume, speed, and incident level in each direction. Pixel-wise mean squared error (MSE) is used to evaluate the performance for ranking the submitted prediction results.

% \begin{figure}[bt]
% \hl{TO BE UPDATED}
% \centering
% \subfloat[{\label{traffic4cast:Berlin}Berlin}]{\includegraphics[width=0.3\linewidth]{pix/traffic4cast/Berlin.png}}

% \subfloat[{\label{traffic4cast:Istanbul}Istanbul}]{\includegraphics[width=0.3\linewidth]{pix/traffic4cast/Istanbul.png}}
% \subfloat[{\label{traffic4cast:Moscow}Moscow}]{\includegraphics[width=0.3\linewidth]{pix/traffic4cast/Moscow.png}}\\
% \caption{\label{traffic4cast:samples} Sample Images from traffic4cast data set for May 1\textsuperscript{st}, 2019. In general, the greener the color, the heavier the road.}
% \end{figure}

\section{Solution}
In this section, we will introduce most of the technical details for our solution along with some experimental results. 

\subsection{Models}
\subsubsection{Choice of Model Architecture}
\begin{figure}[ht]
\centering
\includegraphics[width=\linewidth]{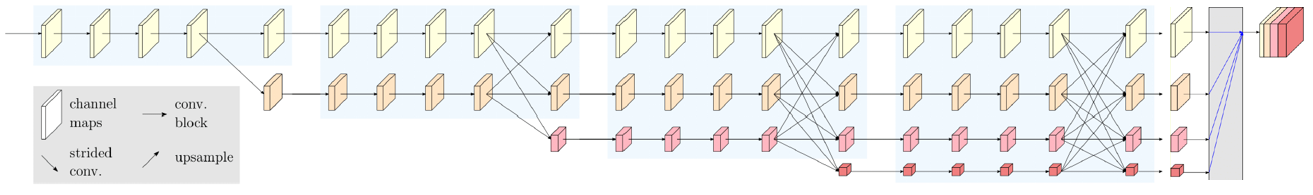}
\caption{\label{img:hrnet}Visualization of HR-NET.}
\end{figure}

In the 2019 Traffic4cast Challenge, was adopted U-NET~\cite{ronneberger2015u}  as the basic model. This year, we introduce HR-NET~\cite{wang2020deep} (see \hyperref[img:hrnet]{Figure}~\ref{img:hrnet}) in the competition, where HR-NET is an advanced network architecture for image segmentation that has demonstrated extraordinary performance in many tasks.

\hyperref[tab:model-comparsion]{Table}~\ref{tab:model-comparsion} shows a comparison of two basic models. We can conclude that, in general, HR-NET performs better than U-NET. Therefore, for the final solution, HR-NET is adopted as our backbone architecture. In the following section, most experiment results are based on HR-NET.

\begin{table}[ht]
\centering
\begin{tabular}{lccc}
\hline
\textbf{Model} & \textbf{Berlin}&\textbf{Istanbul} &\textbf{Moscow} \\
\hline
UNET-EfficientNetB3&1.3175e-3&0.9214e-3&\textbf{1.3701e-3}\\
HRNET-W18&1.2937e-3&\textbf{0.9100e-3}&1.3705e-3\\
HRNET-W48&\textbf{1.2919e-3}&-&-\\
\hline
\end{tabular}
\caption{\label{tab:model-comparsion}Comparison of models in terms of the MSE on the validation set of each city using same features. }
\end{table}

\subsubsection{Choice of Hidden Layer Activation Function}

The commonly used ReLU activation function is not the optimal activation function for many tasks, hence we experimented with some alternatives. Given limited RAM, we focused on in-place activation functions only. \hyperref[tab:activation]{Table}~\ref{tab:activation} lists the hidden activation functions we tested. ELU performed surprisingly better than any other activation function. Therefore, we used ELU as the hidden layer activation function in most places of the final solution.

\begin{table}[ht]
\centering
\begin{tabular}{lccc}
\hline
\textbf{Type} & \textbf{Berlin}&\textbf{Istanbul} &\textbf{Moscow} \\
\hline
ReLU&1.2980e-3&-&-\\
ELU&\textbf{1.2951e-3}&-&-\\
ReLU6&1.2980e-3&-&-\\
LeakyReLU&1.2982e-3&-&-\\
\hline
\end{tabular}
\caption{\label{tab:activation}Comparison of the hidden layer activation functions in terms of the MSE on the validation set of each city using the same inputs.}
\end{table}

\subsection{Features}
Essentially, this task is a time-series prediction problem. For one-dimensional time-series prediction, a typical solution is to transform the problem into a supervised learning problem through feature engineering.  In addition to the given $12\times9$ spatiotemporal channels and nine fixed spatial channels as the inputs, extra features ( \textit{e.g.}, periodic features, and holiday features) were also incorporated and valued. In summary, considering the decrease in MSE loss, we introduce three different types of features in this section.

\subsubsection{Periodic features}

The similarity between traffic states of two different days can be attributed to the periodic characteristics of traffic states, which typically repeat every 24 hours. In total, we use $10\times8$ daily average statistics from $\{D-7\} \bigcup[D-3, D-1]\bigcup [D+1, D+3]\bigcup\{D+7\}$, where $D$ is the predicted day. Since the training set and test set are spitted based on time, we cannot obtain full observation of data in one day during testing. To relieve the gap between training and testing, we randomly sampled 1-4 periods (12-48 time steps) in one day and used it to estimate the average daily statistics. 

\begin{table}[ht]
\centering
\begin{tabular}{lccc}
\hline
\textbf{Type} & \textbf{Berlin}&\textbf{Istanbul} &\textbf{Moscow} \\
\hline
Without Periodic Features&1.3040e-3&0.9173e-3&1.3807e-3\\
With Periodic Features&\textbf{1.2980e-3}	&\textbf{0.9165e-3}&\textbf{1.3764e-3}\\
\hline
\end{tabular}
\caption{\label{tab:holiday}Validity of periodic features based on the MSE on the validation set of each city using the same inputs.}
\end{table}

\subsubsection{Time, Weekday and Holiday Features}
Time, weekday, and holiday features are definitely useful in traffic flow prediction tasks. We used a two-dimensional vector to represent time in one day by projecting $[0,287]$ to a unit circle. We used a one-hot vector to represent weekday and a Boolean value to represent the holiday.  Holiday information was obtained from \url{www.officeholidays.com}. \hyperref[tab:holiday]{Table}~\ref{tab:holiday} shows strong validity of holiday features, which is in line with our expectations.

\begin{table}[ht]
\centering
\begin{tabular}{lccc}
\hline
\textbf{Type} & \textbf{Berlin}&\textbf{Istanbul} &\textbf{Moscow} \\
\hline
Without Holiday Features&1.2980e-3&0.9165e-3&1.3764e-3\\
With Holiday Features&\textbf{1.2937e-3}	&\textbf{0.9100e-3}&\textbf{1.3705e-3}\\
\hline
\end{tabular}
\caption{\label{tab:holiday}Validity of holiday features based on the MSE on the validation set of each city using same inputs. Note that features and dataset used here might not be the same as other tables in this paper.}
\end{table}

\subsubsection{Geo-Embedding features}
The locality is a common assumption in image segmentation and classification task that the object should be identical irrespective of its position in the image. This assumption is not valid for spatiotemporal data, as each pixel in the spatiotemporal data represents a region in the physical world, which has its inherent attributes. To adapt semantic segmentation models to spatiotemporal data, it is necessary to develop a technique to learn the inherent attributes of each pixel. In our previous studies~\cite{liu2020building}, we used embedding technique to generate regional 'personalized' temporal information and feed it into the convolutional neural network. In this competition, we further propose the method of geo-embedding to learn the inherent attributes of each location (\textit{i.e.},pixel). We concatenate a learnable tensor $C\times N \times M$ to each input and optimize it using the model.  For the implementation, we use the \textsc{nn.Embedding} in Torch by assigning a different ID to each pixel since \textsc{nn.Embedding} has options of norms to parameters. \hyperref[tab:embed]{Table}~\ref{tab:embed} lists the effects of geo-embedding features, and the contribution of geo-embedding features can be observed.

\begin{table}[ht]
\centering
\begin{tabular}{lccc}
\hline
\textbf{Type} & \textbf{Berlin}&\textbf{Istanbul} &\textbf{Moscow} \\
\hline
Without Embedding Features&1.2919e-3&-&-\\
With Embedding Features&\textbf{1.2913e-3}&-&-\\
\hline
\end{tabular}
\caption{\label{tab:embed}Validity of embedding features based on the MSE on the validation set of each city using the same inputs. }
\end{table}

\subsection{Training}
Due to device issues, most of our models were trained on a mini-batch size of $3\times4$ with $3\times$ 2080Ti GPU.  We typically trained each model with 15 epochs with an initial learning rate of 0.01 and a linear learning rate decay. It typically took 16-20 hours to train a model. We also included \textsc{SyncBatchNorm} to stabilize and speed up the training. In this section, we will introduce some options during training that potentially contribute to model performance.

\subsubsection{Choice of Optimizer}
SGD and ADAM~\cite{diederik2015} are commonly used to optimize the model. In most conditions, \textsc{SGD} is slower but theoretically guarantees to convert, while ADAM is slightly faster, but may not guarantee to convert. Recently, other self-adaptive optimizers have also become popular,  e.g., LAMB~\cite{You2020Large}.  It should be noted that, we did not optimize the learning rate for SGD; hence it might be possible, but less likely, that after a careful design of the initial learning rate for SGD, the result will be comparable to the self-adaptive optimizer in the same training time. We compare several optimizers in \hyperref[tab:optim]{Table}~\ref{tab:optim}, and LAMB appears to be the best optimizer for this task.   

\begin{table}[ht]
\centering
\begin{tabular}{lccc}
\hline
\textbf{Type} & \textbf{Berlin}&\textbf{Istanbul} &\textbf{Moscow} \\
\hline
SGD&1.8271e-3&-&-\\
ADAM &1.3067e-3&-&-\\
ADAMW &1.3042e-3&-&-\\
LAMB &\textbf{1.3040e-3}&-&-\\
\hline
\end{tabular}
\caption{\label{tab:optim}Comparison of optimizers based on the MSE score on the validation set of each city using the same inputs. The number of epochs is set to 15 and initial learning rate is set to 0.01.}
\end{table}

\subsubsection{Warm-up}

Learning rate warm-up has been used for many NLP tasks. \hyperref[tab:lr]{Table}~\ref{tab:lr} lists results of our examination of the effects of warm-up learning rate. By fixing the training epochs and learning rate, using warm-up can generate far better results. 

\begin{table}[ht]
\centering
\begin{tabular}{lccc}
\hline
\textbf{Type} & \textbf{Berlin}&\textbf{Istanbul} &\textbf{Moscow} \\
\hline
Without warm-up&1.6280e-3&-&-\\
With warm-up &\textbf{1.3067e-3}&-&-\\
\hline
\end{tabular}
\caption{\label{tab:lr} Validity of warm up based on the MSE on the validation set of each city using the same inputs. Note that  features and dataset used here might not be the same asin  other tables in this paper.}
\end{table}

\subsubsection{Inclusion of Validation Set into Training Process}

After two weeks of experiments and submissions, we found that the offline validation set MSE score and the online MSE score were consistent, even though we greedily selected the best model parameter based on validation MSE. This phenomenon gave us hint to include the validation data in the training process. In the final solution, half of the models are trained by both the training set and the validation set.

% \section{Online Submission Results}
% \hyperref[tab:online]{Table}~\ref{tab:online} lists all scores of our online submission. Our best single model is 10-24-11 (1.1761e-3), which uses all the above best methods and features in each section, and our best submission is 11-02-19, which is an ensemble of seven models, and those models are different in architectures, features, and data-set.

% \begin{table}[ht]
% \centering
% \begin{tabular}{lll}
% \hline
% \textbf{Name} & \textbf{Score} & \textbf{Configuration}\\
% \hline
% 09-29-10&1.3600e-3&v1-hrnet-base-lr-001\\
% 09-30-10&1.3334e-3&v1-hrnet-base-lr-001 and v1-hrnet-base-lr-001-all\\
% 10-01-19&1.2019e-3&v2-hrnet-base\\
% 10-05-22&1.1980e-3&v3-hrnet-10-day\\
% 10-08-03&1.1917e-3&v4-hrnet-base\\
% 10-12-02&1.1846e-3&v4-hrnet-base and v4-unet-base\\
% 10-12-08&1.1911e-3&v4-hrnet-base-include-valid\\
% 10-12-10&1.1823e-3&v4-hrnet-base and v4-unet-base\\
% 10-13-06&1.1814e-3&v4-hrnet-include-valid and v4-unet-base\\
% 10-14-05&1.1805e-3&v4-hrnet-include-valid and v4-unet-include-valid\\
% 10-15-00&1.1735e-3&3 models\\
% 10-18-05&\textbf{1.1688e-3}&4 models\\
% 10-20-04&1.1833e-3&v4-hrnet-w48-geo-embed-include-valid-10-more\\
% 10-24-11&\textbf{1.1761e-3}&v4-hrnet-w48-geo-embed-include-valid\\
% ...\\
% 11-02-19&\textbf{1.1667e-3}&7 models\\ 
% \hline
% \end{tabular}
% \caption{\label{tab:online}History of Online Submission.}
% \end{table}

\section{Conclusion}
In this paper, we conducted systematic research on large-scale spatiotemporal traffic prediction. The model structure is designed to accommodate spatiotemporal prediction based on the image segmentation model. Targeting spatiotemporal properties of traffic data, we proposed several new methods, including geo-embedding, and explored the details and tricks in model training. It should be noted that one assumption of the current dynamic traffic state prediction model is that no significant external influences exist.However, in reality, such as when large events are held, the model's predictive performance is drastically reduced. For future research, we will investigate effective traffic prediction under strong external influences.

\section*{About Us}
\textbf{Fanyou Wu} is pursuing a Ph.D. degree in Forestry and Natural Resources Department, Purdue University, West Lafayette, USA. He has gained a wealth of experience in the theory of interdisciplinary applications of machine learning techniques and has won many championships in AI competitions organized by leading international AI conferences or research institutes, including the championship of JDD (2019), championship of IJCAI-Adversarial AI Challenge (2019), and championship of KDD Cup (2020).

\textbf{Yang Liu} is pursuing a Ph.D. degree in Transportation Engineering in the School of Transportation at Southeast University, Nanjing, China. He has rich experience in artificial intelligence applications, covering diverse fields from spectral classification for LAMOST to local climate zone classification for Sentinel-1, and from short video recommendation for TikTok users to travel mode recommendation for travelers. He has won three championships of Alibaba’s Tianchi Algorithm Competition, championship of IJCAI-Adversarial AI Challenge (2019), and championship of KDD Cup (2020).
\clearpage
\bibliography{main}
\end{document}